\documentclass[pdf-a,balance,colorlinks,upint,subscriptcorrection,varvw,mathalfa=cal=boondoxo, spanish,french,vietnamese,russian,greek]{asmeconf}

\hypersetup{%
	pdfauthor={Georg Siedel},									  
	pdftitle={A Practical Approach to Evaluating the Adversarial Distance for Machine Learning Classifiers},                  
	pdfkeywords={Robustness, Adversarial Distance, AI Safety},
	pdfsubject = {IMECE paper on adversarial distance for classifiers},			  
	pdflicenseurl={https://ctan.org/pkg/asmeconf},
}

\usepackage{graphicx, amsmath, algorithm, algpseudocode}

\usepackage{booktabs} 
\usepackage{multirow} 
\begin{document}

\ConfName{Proceedings of the ASME 2024\linebreak International Mechanical Engineering Congress and Exposition}
\ConfAcronym{IMECE2024}
\ConfDate{November 17--21, 2024} 
\ConfCity{Portland, OR} 
\PaperNo{IMECE2024-145280}

\title{A Practical Approach to Evaluating the Adversarial Distance for Machine Learning Classifiers}

\SetAuthors{%
	Georg Siedel\affil{1}\affil{2}\JointFirstAuthor\CorrespondingAuthor{siedel.georg@baua.bund.de, ekagrag99@gmail.com}, 
	Ekagra Gupta\affil{2}\JointFirstAuthor\CorrespondingAuthor{},
	Andrey Morozov\affil{2}
}

\SetAffiliation{1}{Federal Institute for Occupational Safety and Health (BAuA), Dresden, Germany}
\SetAffiliation{2}{University of Stuttgart, Germany}

\maketitle


\begin{abstract}
Robustness is critical for machine learning (ML) classifiers to ensure consistent performance in real-world applications where models may encounter corrupted or adversarial inputs. In particular, assessing the robustness of classifiers to adversarial inputs is essential to protect systems from vulnerabilities and thus ensure safety in use. However, methods to accurately compute adversarial robustness have been challenging for complex ML models and high-dimensional data. Furthermore, evaluations typically measure adversarial accuracy on specific attack budgets, limiting the informative value of the resulting metrics.
This paper investigates the estimation of the more informative adversarial distance using iterative adversarial attacks and a certification approach. Combined, the methods provide a comprehensive evaluation of adversarial robustness by computing estimates for the upper and lower bounds of the adversarial distance. We present visualisations and ablation studies that provide insights into how this evaluation method should be applied and parameterised. We find that our adversarial attack approach is effective compared to related implementations, while the certification method falls short of expectations. The approach in this paper should encourage a more informative way of evaluating the adversarial robustness of ML classifiers.
\end{abstract}


\section{Introduction}

\begin{figure}[t]
    \centering
    \begin{subfigure}{0.55\linewidth} 
        \centering
        \includegraphics[width=\linewidth,trim={0 0 3cm 3.5cm},clip]{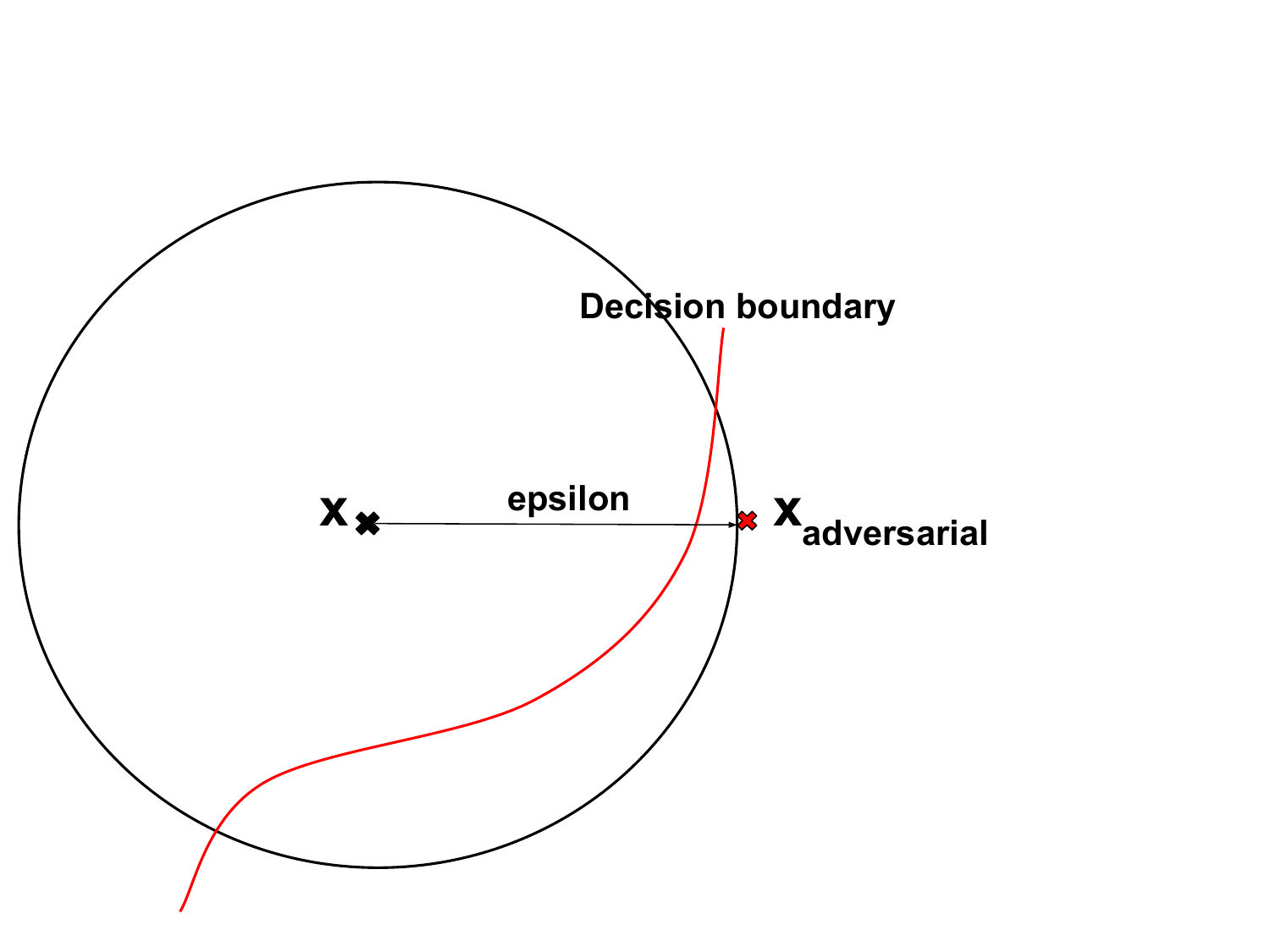}
        \caption{Non-iterative Attack}
        \label{fig:noniterative}
    \end{subfigure}\hfill 
    \begin{subfigure}{0.45\linewidth}
        \centering
        \includegraphics[width=\linewidth,trim={0 0 7cm 3.5cm},clip]{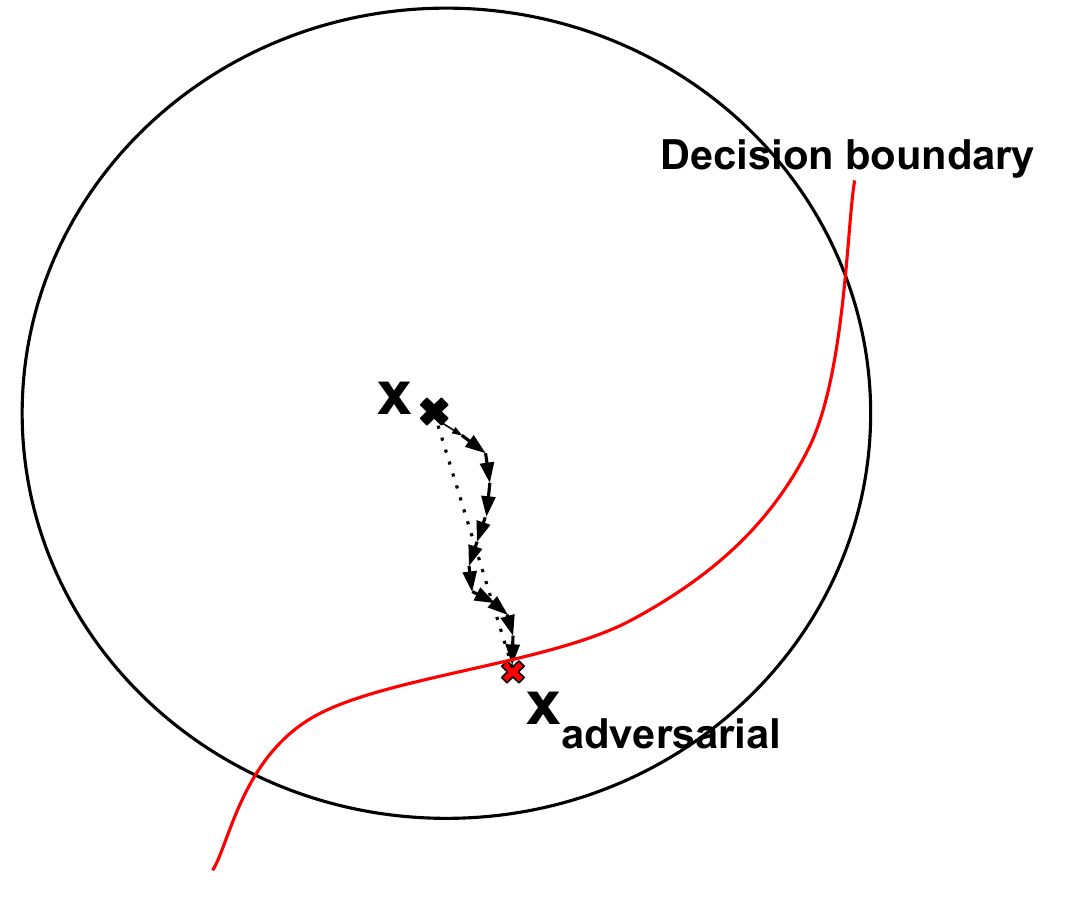}
        \caption{Iterative Attack} 
        \label{fig:iterative}
    \end{subfigure}
    \caption{A one-step attack and an iterative attack on the model with the red decision boundary visualize the difference between (a) finding some adversarial perturbation $x_{adversarial}$ using the full attack budget epsilon and (b) finding a good estimate for the minimal adversarial distance between $x_{adversarial}$ and $x$. In case (a), only a discrete 0-or-1 success/failure rate (adversarial risk/accuracy) of the attack can be reported, while (b) allows for a continuous robustness assessment.}
    \label{fig:adversarial-attacks}
\end{figure}

Robustness is a prerequisite for the safe and secure implementation of ML in high-risk applications. The recently published proposal for a horizontal European regulation on artificial intelligence (AI), the "AI Act", explicitly addresses robustness as an essential property \cite{AIAct2024}. High-risk AI on the European market will be required to meet certain levels of accuracy, robustness and cybersecurity to ensure a trustworthy and safe application according to Article 15 \cite{AIAct2024}. Robustness is the ability of an AI module to cope with erroneous, noisy, unknown, or adversarially constructed input data \cite{DINSPEC92001-2}. Specifically, adversarial robustness (AR) refers to robustness against "attempts to deceive the AI module by means of carefully chosen harmful input" \cite{DINSPEC92001-2}. AR can be interpreted as a worst-case scenario for robustness under the threat of an attacker. Thus, AR is required by two of the above-mentioned requirements of the AI Act at the same time: cybersecurity and robustness. The AI Act encourages the development of benchmarks and  measurement methodologies "to address the technical aspects of how to measure the appropriate levels of accuracy, robustness and cybersecurity" \cite{AIAct2024}. However, despite ongoing research and discussion as required by the AI Act \cite{guo2023comprehensive, villegas2024evaluating}, there exists a gap in the development of standardised, widely applicable and meaningful measurement methodologies for assessing the robustness of ML systems \cite{STDReq}.

Current research typically evaluates adversarial robustness based on accuracy, which requires the prior definition of an attack budget and solely reports a discrete 0-or-1 success rate. Based on the above motivation to develop measures of adversarial robustness, this paper focuses on a less common but more informative metric: (minimal) Adversarial Distance (AD), for which we compute estimates for upper and lower bounds (see Figure \ref{fig:adversarial-attacks}). We find that many popular methods, or their implementations, are ill-suited for accurately estimating adversarial distance. It appears that while the techniques for AD evaluation are available, their correct application using a popular software package is not straightforward.

Our contributions to this problem are as follows: 
\begin{enumerate}
    \item We propose an efficient attack algorithm for estimating adversarial distance, which serves as a baseline measure.
    \item We combine this baseline with other appropriate approaches in an "estimation ensemble", theoretically bounding the true adversarial distance above and below,
    \item We perform experiments on the parameterisation, effectiveness and computational efficiency of the proposed estimation methods, evaluating two differently robust models on an image classification task\footnote{Code available: \href{https://github.com/georgsiedel/adversarial-distance-estimation}{github.com/georgsiedel/adversarial-distance-estimation}}.
\end{enumerate}


\section{Preliminaries}

Given a classifier $f$, an adversarial perturbation is typically defined as the minimal perturbation $r$ that changes the estimated label $f(x)$:
    \begin{align}
        \Delta_{\text{adv}}(x, f) = \min_{r \in \mathbb{R}^d} \| r \|_p \quad \text{subject to} \quad f(x) \neq f(x + r)     
        \label{eq:adv-pert}
    \end{align}

where ${x \in \mathbb{R}^d}$ is a data point and $p$ is a norm to quantify the perturbation \cite{ren2020adversarial, fawzi2018analysis, moosavi2016deepfool}. Although adversarial perturbations have been defined according to other and more general distance measures \cite{carlini2017towards}, $L_2$ and $L_\infty$ are the most common norms used for adversarial robustness evaluation \cite{fawzi2018analysis}.
$\Delta_{\text{adv}}(x, f)$ is then called the robustness of $f$ at $x$. With  $\mathbb{E}_\mu$ defined as the expected value over all $x$ sampled from distribution $\mu$, the overall robustness of $f$ is defined as:

    \begin{align}
\rho_{\text{adv}}(f) = \mathbb{E}_\mu(\Delta_{\text{adv}}(x, f)).
    \label{eq:adv-robustness}
    \end{align}

In words, it is defined as the average norm of the minimal perturbation required to change the predictions of $f$ across all data points \cite{fawzi2018analysis}. For this reason, this measure is also referred to as the (mean) adversarial distance \cite{gilmer2018adversarial}. 

Research on adversarial robustness evaluation has focused on methods that find such adversarial perturbations through attacks, while solving \eqref{eq:adv-pert} directly is a less popular research direction in machine learning \cite{zhao2018admm}. Adversarial examples found by an attack are always a guaranteed upper bound on the true minimal adversarial perturbation, no matter how tight they are.

For the purpose of benchmarking models and defences on datasets, most adversarial attack and certification approaches define a specific attack budget $\epsilon$ instead of trying to find the minimal adversarial distance \cite{croce2020robustbench}. This attack budget bounds the norm of the maximum allowed perturbation. The adversarial robustness is then evaluated according to the probability of correct classification using this given attack budget:

    \begin{align}
P_{\text{adv}}(f) = \mathbb{E}_\mu(f(x')=y) \quad \text{for all} \quad x' \in \mathcal{B}(x, \epsilon).
    \label{eq:adv-accuracy}
    \end{align}

where $\mathcal{B}(x, \epsilon)$ is a norm ball defining the given attack budget and $y$ is the true label for $x$. $P_{\text{adv}}(f)$ is typically called "adversarial accuracy" or "astuteness" \cite{yang2020closer}, while "adversarial risk" is its opposite term. Adversarial accuracy can be estimated with a selected attack and attack budget on a given test dataset. 

The evaluation approach \eqref{eq:adv-accuracy} is quite different from \eqref{eq:adv-robustness}. For adversarial accuracy evaluation, attacks can always fully exploit their attack budget. Reporting adversarial accuracy is well comparable for a fixed attack type and attack budget. However, the metric is less informative than \eqref{eq:adv-robustness}, as it simply sums up the successes or failures of an attack as 0 or 1. For example, consider two models that could behave very differently under different attacks or in real-world applications: The first is a classifier that is robust just below the given attack budget on all tested data points, and the second is not robust at all. Both would obtain an adversarial accuracy of $0\%$ according to \eqref{eq:adv-accuracy}, but different mean adversarial distances according to \eqref{eq:adv-robustness}, providing a more nuanced and more informative assessment of the models adversarial robustness. Despite its advantages, adversarial distance is only rarely reported in basic research such as \cite{szegedy2013intriguing} and is absent from the evaluations of most adversarial defense methods.

\section{Related Work}
\subsection{Adversarial Distance Calculation through Attacks}

There exist approaches that pave the way for estimating the upper bound adversarial distance according to \eqref{eq:adv-robustness} by focusing on generating minimal adversarial perturbations instead of using the entire attack budget. 

Single-step adversarial attacks are generally not well suited for estimating tight minimal adversarial perturbations, as they generate a deceptive example in one step. One such attack is the Fast Gradient Sign Method (FGSM) \cite{goodfellow2014explaining}, where a single perturbation is applied to the original input along the gradient of the classifier. In a non-iterative setting, the attack will use its full attack budget $\epsilon$, producing suboptimally large minimal perturbation estimates (see Figure \ref{fig:noniterative}).

Iterative attacks, on the other hand, incrementally alter the input while staying within the predefined attack budget. As shown in Figure \ref{fig:iterative}, these methods push the perturbation towards the decision boundary. Popular iterative adversarial attacks include Projected Gradient Descent (PGD) \cite{madry2017towards} and Basic Iterative Method (BIM) \cite{kurakin2018adversarial}. Iterative methods do not need to use their entire attack budget epsilon, as they could potentially early stop once they have successfully changed the class. The smaller the step size, the more applicable they become for estimating tight minimal perturbations according to \eqref{eq:adv-robustness}.

The authors of \cite{moosavi2016deepfool} emphasise adversarial distance as a refined measure of robustness beyond adversarial accuracy. They propose an adversarial attack called DeepFool for $L_{2}$ norm, which estimates a tight minimal perturbation using an iterative attack that stops when the prediction changes.

A related approach to DeepFool is NewtonFool \cite{jang2017objective}. This method also aims to create minimal perturbations, meaning that they are in principle suitable for adversarial distance estimation.

The authors of \cite{carlini2017towards} propose the Carlini-Wagner (CW) attack that uses an iterative search strategy to find close adversarial perturbations. Their method is reported to be particularly effective for the $L_{2}$ norm. The authors are also among the few to report adversarial distance for successful attacks in their experiments.

\subsection{Robustness Certification}
In contrast to upper bound adversarial distance calculation through attacks, certification methods\footnote{Note that certification of classifier robustness is unrelated to efforts of official institutions and authorities that may certify products.} aim at identifying a lower bound on the true minimal adversarial perturbation, namely a norm distance for which no adversarial perturbation exists. One line of research is on using formal mathematical proofs to verify robustness \cite{meng2022adversarial, liu2021algorithms}. 

Another is robust defense approaches that also claim theoretically guaranteed certified robustness. Sometimes, such papers plot guaranteed adversarial accuracy over various sizes of adversarial perturbations \cite{cohen2019certified, zhu2022adversarially}. When evaluated for a sufficiently continuous set of perturbation sizes, such plots can be interpreted as an estimate of the cumulative distribution function for the lower bounds of the adversarial distances of all data inputs.

\subsection{Clever Score Metric}

The CLEVER (\textbf{C}ross \textbf{L}ipschitz \textbf{E}xtreme \textbf{V}alue for n\textbf{E}twork \textbf{R}obustness) score \cite{weng2018evaluating} is a popular and scalable robustness certification approach that estimates a lower bound on adversarial distance. It represents an attack-agnostic metric applicable to all neural network classifiers. It is based on the principle of Lipschitz continuity \cite{eriksson2004lipschitz}. 

\begin{figure}[t]
\centering
\includegraphics[width=\linewidth]{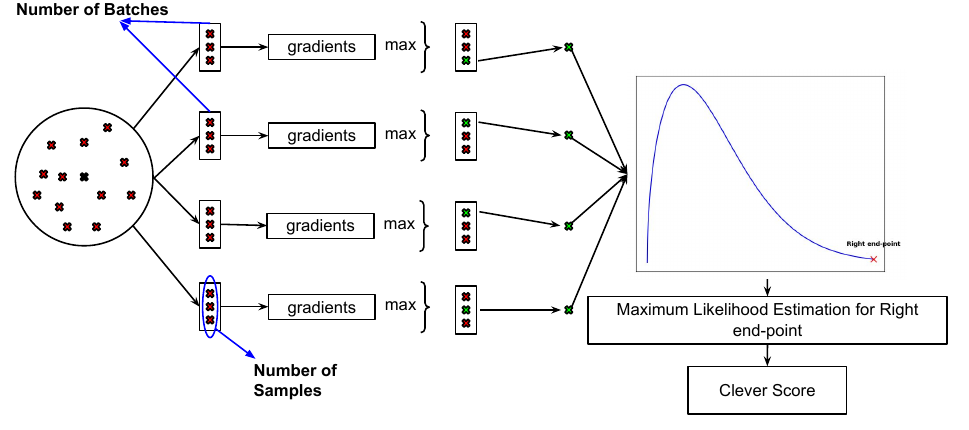}\\[-2pt]
\caption{Working principle of CLEVER score estimation}
\label{fig:clever_flowchart}
\end{figure}

The CLEVER method uniformly generates an additional number of samples in a specified $p$-norm neighbourhood around the original input, as shown in Figure \ref{fig:clever_flowchart}. It then assigns the samples to a specified number of batches and calculates the maximum norm of the local gradients of the samples in the batch. These maximum gradients follow a reverse Weibull distribution according to the principles of extreme value theory \cite{smith1990extreme}. From this distribution, the maximum gradient, which represents the cross-Lipschitz constant, can be estimated as the finite right tail of the Weibull distribution using maximum likelihood estimation. The cross-Lipschitz constant can be used to derive a robustness lower bound for the data point, below which the class cannot be changed, implying adversarial robustness.

It should be noted that CLEVER is an estimation based on statistical sampling and does not provide guarantees.

\subsection{Robustness Toolbox}

There are several popular libraries that implement approaches for adversarial attack, defence, and robustness estimation. One such open source library is the Adversarial Robustness Toolbox (ART) \cite{nicolae2018adversarial}, implemented in Python. 

ART provides an implementation of CLEVER and several adversarial attacks, including those mentioned above as well as HopSkipJump (HSJ) \cite{chen2020hopskipjumpattack} or ElasticNet (EAD) \cite{chen2018ead}. On paper then, the ART toolbox provides all the means to compute upper and lower bounds of \eqref{eq:adv-robustness}.

However, we show in section \ref{section:results} that many attacks, or their implementation in ART, are ill-suited for accurate adversarial distance estimation for one of several reasons. Some iterative approaches do not provide sufficiently tight perturbations, such as the implementation of DeepFool. Other iterative methods, such as the implementation of all PGD variants, do not stop at class change, but use their full attack budget $\epsilon$ instead. ART even provides the metric "Empirical Robustness", which implements a wrapper for some iterative attacks to stop them early in order to estimate the adversarial distance. However, our results show that this metric only works for an iterative FGSM attack, and the resulting perturbation is not competitively tight.

Looking back at the state of the art, we find that calculating upper bounds of adversarial distance in particular is relatively uncommon in the literature. It is no wonder then that implementations for practitioners in a popular package such as ART do not work effectively.

\section{Proposed Algorithm}

Based on our unsatisfactory findings of the minimal adversarial distance calculation in a common robustness toolbox such as ART, we propose a simple approach to adversarial distance calculation. Our method, described in Algorithm \eqref{alg:early_stopping}, can use any attack that returns an intermediate result even if no adversarial example is found. The attack generates a perturbed image $x_{adv}$ in norm $p$ using the input image $x$ and the specific attack parameter $\epsilon_{step}$. The algorithm repeats this generation up to $max\_iters$ times. After each iteration, it checks whether $f(x_{adv})$ diverges from $y$, which indicates a successful adversarial attack. An early stopping mechanism stops the attack in this case, allowing the Algorithm \eqref{alg:early_stopping} to estimate the minimum adversarial perturbation of the image $x$.
It uses an early stopping function to extract the tightest possible adversarial perturbation for this attack.

Note that the principle of this early stopping function has been described in \cite{moosavi2016deepfool} for the DeepFool attack. However, any attack that returns intermediate results after each iteration would be possible (e.g. FGSM, NewtonFool, DeepFool). In this study, we use a simple 1-step PGD attack on the \eqref{alg:early_stopping} algorithm. While this recombination seems obvious, we will see in the experiments that ART does not provide a functionality that leads to the same results.

\begin{algorithm}[t]
\caption{Adversarial Attack with Early Stopping}
\label{alg:early_stopping}
\begin{algorithmic}[1]
\State \textbf{Input:} classifier $f$, input image $x$, true label $y$, step perturbation $\epsilon_{step}$, maximum iterations $max\_iters$, norm $p$, attack_type $attack$
\State \textbf{Output:} $x_{adv}$

\State \textbf{Initialize} $label\_flipped$ $\gets$ $False$, $\epsilon=max\_iters\cdot\epsilon_{step}$

\For{$i$ in range($max\_iters$)}
    \State $x_{adv}$ $\gets$ $attack.generate(f, x, y, \epsilon, \epsilon_{step},\newline\hspace*{1.81in} max\_iters, p)$
    \State $predicted_{adv}$ $\gets$ $f(x_{adv})$

    \State $label\_flipped \gets predicted_{adv} \neq y$
    \If{$label\_flipped$}
        \State \textbf{break}
    \EndIf
    \State $x \gets x_{adv}$
\EndFor

\State \textbf{return} $x_{adv}$
\end{algorithmic}
\end{algorithm}

Algorithm \eqref{alg:combined} extends our approach to a comprehensive evaluation of the adversarial distance of a classifier. It performs two adversarial attacks on each image in the test set, the first using Algorithm \eqref{alg:early_stopping}, the second being a selected second attack that estimates tight minimal perturbations. The choice of the second attack depends on the norm $p$ and is described in section \ref{section:results}. For both attacks, Algorithm \eqref{alg:combined} computes the adversarial distance between the original image $x$ and its adversarial example $x_{adv}$ according to the norm $p$. The algorithm selects the smaller of the two distances for each image, identifying it as the most effective minimal perturbation. Algorithm \eqref{alg:early_stopping} serves as a baseline, while the second, norm-specific attack may produce even tighter minimal perturbations. From the adversarial distances of all data points, the maximum perturbation distance is identified as $radius^{max}_p$. This value is used as the sampling radius for calculating the CLEVER score for the same classifier $f$. The algorithm then outputs adversarial distances and corresponding CLEVER scores for all data points, ordered by increasing size of the adversarial distances. Overall, Algorithm \eqref{alg:combined} provides both upper and lower bounds on the adversarial distance for the classifier and the given data points. The real value should lie between the adversarial distance and the CLEVER score, although only the adversarial distance is a guaranteed and trustworthy upper bound.

\begin{algorithm}[t]
\caption{Upper and Lower bound for Adversarial Attacks}
\label{alg:combined}
\begin{algorithmic}[2]
\State \textbf{Input:} Same as Algorithm \eqref{alg:early_stopping}, $max\_iters_{2}$
\State \textbf{Output:} adversaralDistances, cleverScores
\State \textbf{Initialize} $correctPrediction_1, correctPrediction_2, \newline totalLabels$ with $0$, \State$attack(attack\_type_2, max\_iters_2)\to second\_attack$
\For{each image $x$ in $x_{test}$}
\State $x^1_{adv} \gets$ \Call{Algorithm \eqref{alg:early_stopping}}{$max\_iters=1$}
\State $distance^1_p \gets \Call{ComputeDistance}{x, x^1_{adv}, p}$
\hspace{\algorithmicindent}
\State $x^2_{adv} \gets second\_attack.generate(f, x, y)$
\State $distance^2_p \gets \Call{ComputeDistance}{x, x^2_{adv}, p}$
\Statex
\State $adversarialDistances$.append(\newline\hspace*{0.75in}
            \Call{min}{$distance^2_p$, $distance^1_p$})
\EndFor
\State $radius^{max}_p \gets max(adversarialDistances)$
\Statex
\State Perform clever score calculation:
\For{each image $x$ in $x_{test}$}
\State$\Call{ComputeCleverScore}{f, x, y, radius^{max}_p\newline\hspace*{0.25in} nb\_batch, batch\_size} \to cleverScore$
\State$cleverScores.append(cleverScore)$
\EndFor
\Statex
\State $indices \gets \text{argsort}(adversarialDistances)$
\State $sortedCleverScores \gets cleverScores[indices]$
\State \textbf{return} $adversarialDistances, sortedCleverScores$
\end{algorithmic}
\end{algorithm}

\section{Experiments}

We evaluate the effectiveness of our adversarial distance computation with experiments on the CIFAR-10 image classification dataset \cite{krizhevsky2009learning}. 

We compare two pre-trained models according to the Wide Residual Network 28-4 architecture described in \cite{zagoruyko2016wide}. The "standard" model is trained with a typical Pytorch training pipeline using only dropout and standard random crop and flip data augmentations as described in \cite{zagoruyko2016wide}. The "robust" model is additionally trained with a combined set of random data augmentations, including Mixup \cite{zhang2017mixup}, TrivialAugment \cite{muller2021trivialaugment}, and random $p$-norm noise injections \cite{siedel2023investigating}. It is therefore expected to be more robust. Although the model is not trained for adversarial robustness, which \cite{fawzi2018analysis} is a much harder goal to achieve than robustness against random corruptions, the adversarial distance metric should be able to measure the higher robustness of the robust model, even if both models are not adversarially trained. We also compare an "adversarial" model of similar architecture from \cite{Ding2020MMA} loaded from \cite{croce2020robustbench}, which is expected to yield much higher adversarial distance.

The computations for which we report runtime evaluations were performed on the data science platform Kaggle, using a NVIDIA P100 as a GPU.

We evaluated adversarial distances for three common norms:
\begin{enumerate}
    \item $L_{1}$ Distance ($1$-Norm) is defined as:
        \begin{align}
        \|x - x_{adv}\|_1 = \sum_i|x_i - x_{adv_i}|
        \label{eq:L1}
    \end{align}
 $L_{1}$ perturbations make intense changes to few pixels.
    
    \item $L_{2}$ Distance ($2$-Norm / Euclidean distance) is defined as:
        \begin{align}
        \|x - x_{adv}\|_2 = \sqrt{\sum_i(x_i - x_{adv_i})}
        \label{eq:L2}
    \end{align}
    $L_{2}$ perturbations are more evenly spread across all pixels.
    
    \item $L_{\infty}$ Distance ($\infty$ Norm) is defined as: 
        \begin{align}
        \|x - x_{adv}\|_\infty = max_i \cdot |x_i - x_{adv_i}|
        \label{eq:Linf}
    \end{align}
    This is probably the most common adversarial distance metric. This type of attack tends to be the least visible, as it limits the maximum change to any pixel in the image uniformly.

\end{enumerate}

Algorithm \eqref{alg:early_stopping} should change the class on all images, and we initially set $max\_iters = 500$, a value high enough for the standard and robust models. For the adversarial model, this attack budget is not high enough, so it was increased to $max\_iters = 10000$. The overall maximum perturbation $\epsilon$ is then defined by Algorithm \eqref{alg:early_stopping} to be $max\_iters * \epsilon_{step}$. The only important parameter to tune in this algorithm is $\epsilon_{step}$, for which we report sensible parameters in section \ref{section:results}.

We have compared our Algorithm \eqref{alg:early_stopping} with PGD with several potential second attacks. HSJ, EAD and CW do not return intermediate results if they cannot find an adversarial example after one iteration. They are therefore not suitable to be plugged into our \eqref{alg:early_stopping} algorithm, but can still find tight adversarial perturbations on their own. We run them with 40 iterations (100 for the adversarial model), as done in \cite{madry2017towards} for the CW attack, and the runtime is reported with this parameterisation. For DeepFool and NewtonFool it makes no difference whether we plug them into Algorithm \eqref{alg:early_stopping} or let them run independently with the same $\epsilon_{step}$. We also tested the iterative FGSM implementation in ART, which is deliberately designed to evaluate adversarial distance. Finally, we experimented with ART implementations of different PGD variants and AutoAttack \cite{croce2020reliable}. These attacks never stop early when changing a class, even when plugged into our Algorithm \eqref{alg:early_stopping}. They are therefore inapplicable for adversarial distance computation and are not reported upon.

\begin{figure}[t]
\centering\includegraphics[width=\linewidth]{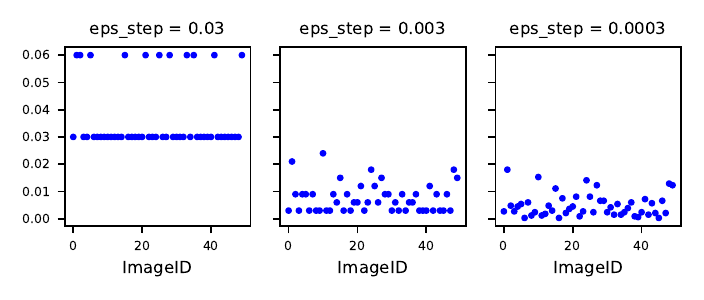}\\[-11pt]
\caption{$L_{\infty}$ Distance of 50 images for different $\epsilon_{step}$ values (standard model)}
\label{fig:clustering-eps}
\end{figure}

\begin{figure}[t]
    \centering    \includegraphics[width=0.9\linewidth,trim={0.4cm 0.48cm 0.4cm 0.1cm},clip]{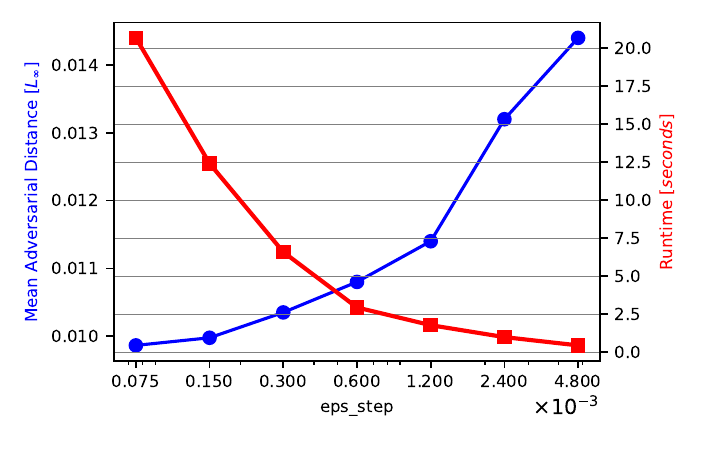}\\[-4pt]
    \caption{Trade-off between tightness of the mean adversarial distance estimation and the computational effort for varying $\epsilon_{step}$}
    \label{fig:tradeoff-curve-eps}
\end{figure}

\section{Results}
In sections \ref{section:results1} and \ref{section:results2} we present results on a reasonable parameterisation of Algorithm \eqref{alg:early_stopping} and the CLEVER computation. In sections \ref{section:results3} to \ref{section:results5} we present results for adversarial distances on the three norms. We justify which attack is effective enough to be used as a second attack and evaluate the results of the CLEVER score lower bound estimation for all norms. 
\label{section:results}

\subsection{Parameterization of Algorithm 1 with PGD}
\label{section:results1}

In a practical framework, $1/255$ represents the smallest real-world step size feasible in the pixel space. However, for a theoretical white-box evaluation of adversarial robustness, this value can be adjusted. Figure \ref{fig:clustering-eps} shows a clustering effect for high values of $\epsilon_{step}$, leading to inaccurate discrete overestimates of adversarial distance and requiring a precise parameterization of $\epsilon_{step}$ for our approach. The smaller $\epsilon_{step}$ becomes, the higher the resolution and the tighter the minimal perturbation is for the approach according to Algorithm \eqref{alg:early_stopping}. 

Figure \ref{fig:tradeoff-curve-eps} shows how the average $L_\infty$ minimal perturbation on 20 selected images becomes smaller as $\epsilon_{step}$ is reduced. At the same time, more steps are required for each image to change the class, increasing the computation time and revealing a trade-off between performance and computation time. Since the overall computation time for Algorithm \eqref{alg:early_stopping} is relatively small, we choose a relatively small step size of $\epsilon_{step} = 0.0003$ for $L_{\infty}$ distance and $0.005$/$0.2$ for $L_2$/$L_1$ respectively. In addition to the trade-off described above, this choice should also take into account the robustness of the image classifier.

\begin{table}[t]
\caption{CLEVER runtimes, scores and error rates for different parameters ($L_2$ norm, 500 images)}
\centering
\setlength\tabcolsep{5.2pt}
\resizebox{1.0\linewidth}{!}{%
\begin{tabular}{@{\extracolsep{\fill}}llccc@{\extracolsep{\fill}}}
\toprule
Model & Samples/Batches & $\text{Time}[h]$& $\overline{\text{CLEVER}}$& Err.[\%]\\
\midrule
\multirow{4}{*}{Standard} & 5/5 & 0.254 & 0.1170 & 37.29 \\  
&20/10 & 0.44   & 0.0997  & 26.27 \\  
&100/50 & 1.91 & 0.0986 & 20.34 \\  
&1024/500 & 18.55  & 0.0899  & 15.25 \\ [6pt]
\multirow{4}{*}{Robust} & 5/5 & 0.215 & 0.3028 & 36.38 \\  
&20/10 & 0.703   & 0.2165  & 25.78 \\  
&100/50 & 3.213 & 0.1738 & 18.50 \\  
&1024/500 & 32.479   & 0.1609  & 17.46 \\ [6pt]
Adversarial & 1024/500 & 31.275   & 0.4016  & 13.88 \\  
\bottomrule
\end{tabular}}
\label{tab:clever-params}
\end{table}

\begin{figure*}[t]
    \centering
    \begin{subfigure}[t]{0.33\linewidth}
        \includegraphics[width=\linewidth,trim={0.41cm 0.2cm 0.8cm 0.35cm},clip]{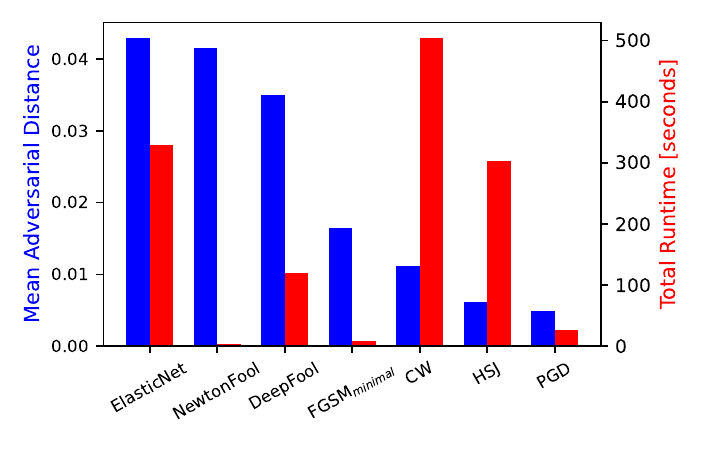}
    \end{subfigure}
    \begin{subfigure}[t]{0.318\linewidth}
        \includegraphics[width=\linewidth,trim={0.8cm 0.2cm 0.8cm 0.1cm},clip]{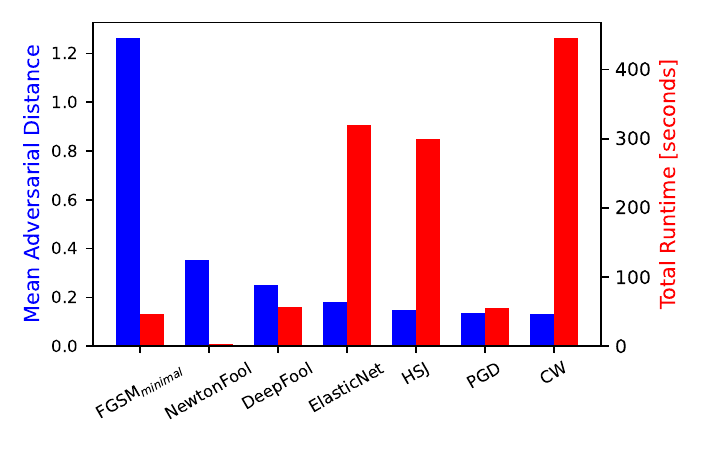}
    \end{subfigure}
    \begin{subfigure}[t]{0.33\linewidth}
        \includegraphics[width=\linewidth,trim={0.8cm 0.2cm 0.43cm 0.1cm},clip]{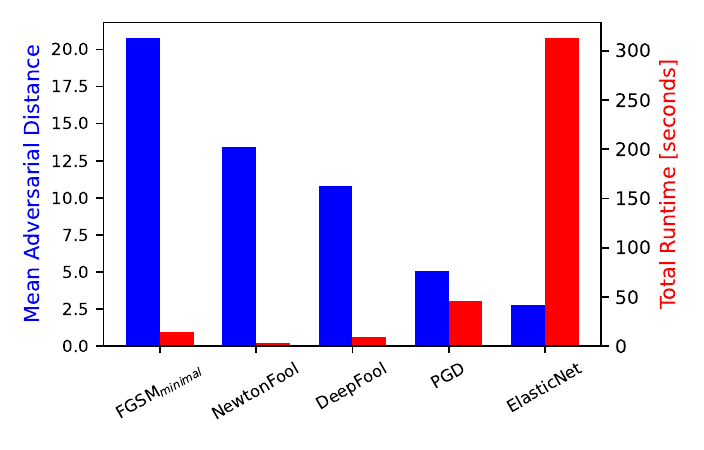}
    \end{subfigure}\\[-6pt]
    \begin{subfigure}[t]{0.33\linewidth}
        \includegraphics[width=\linewidth,trim={0.3cm 0.35cm 0.3cm 0.35cm},clip]{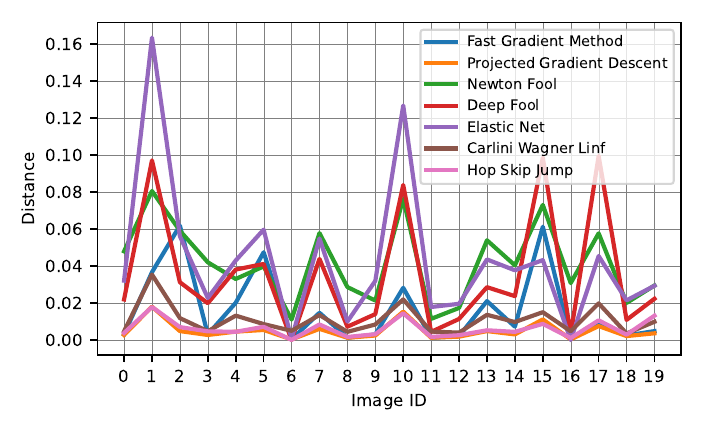}
        \caption{$L_{\infty}$ norm}
        \label{fig:all_attacks_linf}
    \end{subfigure}
    \begin{subfigure}[t]{0.33\linewidth}
        \includegraphics[width=\linewidth,trim={0.3cm 0.35cm 0.3cm 0.25cm},clip]{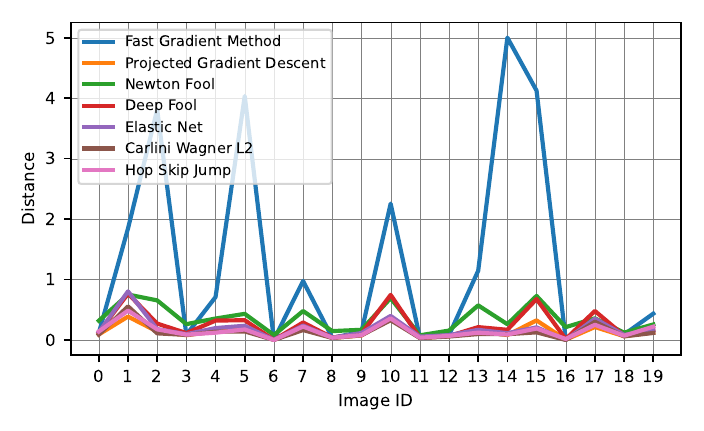}
        \caption{$L_{2}$ norm}
        \label{fig:all_attacks_l2}
    \end{subfigure}
    \begin{subfigure}[t]{0.33\linewidth}
        \includegraphics[width=\linewidth,trim={0.3cm 0.35cm 0.3cm 0.25cm},clip]{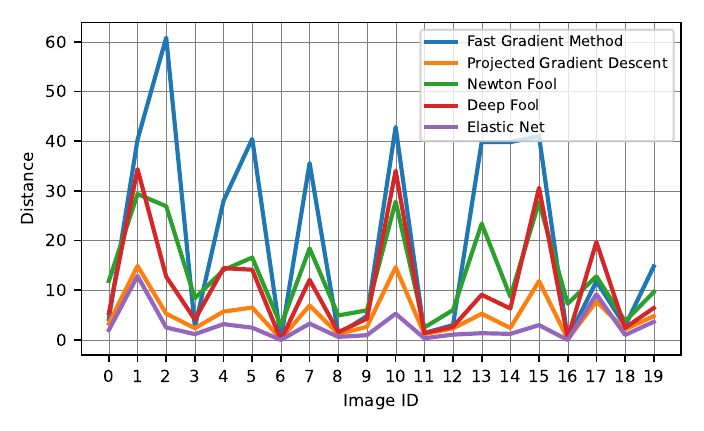}
        \caption{$L_{1}$ norm}
        \label{fig:all_attacks_l1}
    \end{subfigure}\\[-6pt]
    \caption{Comparison of all attacks, standard model, 20 images. Above: Mean Adversarial Distance and total runtime. Below: Image-wise.}
    \label{fig:all_attacks_combined}
\end{figure*}

\subsection{Parameterization of Clever Score}
\label{section:results2}

Table \ref{tab:clever-params} shows, that a small number of samples and batches for CLEVER calculation leads to a larger and, assuming it is a lower bound, less tight mean CLEVER score. Specifically, the lower the number of samples and batches, the higher the proportion of images with a CLEVER score (lower bound) above the adversarial distance (upper bound). For this "Err." ratio of points, CLEVER cannot be a correct lower bound and the metric is not sound. In contrast, setting the number of samples to $1024$ and batchsize of $500$, as parameterised in the original paper \cite{weng2018evaluating}, improves the fraction of correct CLEVER scores. It comes at high computational cost, which scales linearly with the number of samples and the number of classes in the dataset. The decision to use a large sample size must be made keeping in mind that this configuration is the most computationally expensive evaluation of all in this paper. From Table \ref{tab:clever-params} we also find that for $L_2$-norm the error ratio of CLEVER score is lowest for the adversarial trained model. We discuss this phenomenon in more detail in section \ref{section:discussion}.

\subsection{Adversarial Distances in $L_\infty$}
\label{section:results3}

Figure \ref{fig:all_attacks_linf} shows the results of adversarial distance estimation by multiple $L_\infty$-norm attacks on the standard model for 20 images. The EAD, DeepFool and NewtonFool attacks show high variability, often resulting in high perturbation distances. The (iterative) FGSM also fails to consistently induce minimal perturbations, which is surprising since it is the basis of PGD and should work similarly to our Algorithm \eqref{alg:early_stopping} with PGD. CW for $L_\infty$ and HSJ generate solid adversarial perturbations.

The HSJ attack outperforms PGD in perturbing some of the 20 images displayed in \ref{fig:all_attacks_linf}, although not being as effective on average. For $L_\infty$ norm we therefore chose HSJ as the second attack. The advantage of HSJ is that it is a black box method and only requires access to the output of the classifier, not to its internal gradients. However, HSJ requires more time per step compared to PGD, underlining the greater efficiency of PGD. Thus, in situations where time efficiency is a priority, PGD seems to be the more sensible choice in $L_\infty$.

\subsection{Adversarial Distances in $L_2$}
\label{section:results4}

Under the $L_{2}$ norm, we choose CW as the second attack, as it tends to find the tightest adversarial perturbations (see Figure \ref{fig:all_attacks_l2}). CW is expected to produce tight adversarial perturbations in $L_{2}$, as its sophisticated loss function is known to be effective in $L_2$. The Algorithm \eqref{alg:early_stopping} with PGD is still competitive with CW, and produces tighter estimates for some inputs.
Again, the precision of CW as a second attack comes at the cost of its computational time, as it takes about 40 times as long to compute as PGD on the standard model, as shown in Figure \ref{fig:all_attacks_l2}.

\subsection{Adversarial Distances in $L_1$}
\label{section:results5}

Under the $L_{1}$ norm, we choose EAD as the second attack. EAD produces significantly tighter adversarial perturbations compared to Algorithm \eqref{alg:early_stopping} with PGD on almost all data points and on average as can be seen in Figure \ref{fig:all_attacks_l1}. This makes EAD a preferred choice for adversarial distance estimation in $L_1$, despite being computationally more expensive compared to PGD.

\begin{table*}
\caption{Comparative Analysis of Mean Adversarial Distance and Mean Clever Scores Across Standard and Robust Models for Different Norms}\label{tab:adv_cl}
\centering
\begin{tabular*}{\textwidth}{@{\extracolsep{\fill}}llccccc}
\toprule
\multirow{2}{*}{\textbf{Model}} & \multirow{2}{*}{\textbf{Norm}} & \multirow{2}{*}{\textbf{Mean Adversarial Distance (Combined attacks)}} & \multicolumn{4}{c}{\textbf{Mean Clever Score} [Samples/Batches]} \\
\cmidrule(lr){4-7}
& & & [5/5] & [20/10] & [100/50] & [1024/500] \\
\midrule
\multirow{3}{*}{Standard} & $L_1$ & 3.6578 & 0.2855 & 0.1601 & 0.1739 & 0.1775 \\
                           & $L_2$ & 0.1459 & 0.1170 & 0.0997 & 0.0986 & 0.0899 \\
                           & $L_\infty$ & 0.0052 & 0.0053 & 0.0049 & 0.0045 & 0.0042 \\[6pt]
\multirow{3}{*}{Robust}   & $L_1$ & 7.6676 & 4.1041 & 3.1818 & 2.4529 & 2.2521 \\
                           & $L_2$ & 0.3054 & 0.3028 & 0.2165 & 0.1738 & 0.1609 \\
                           & $L_\infty$ & 0.0102 & 0.0169 & 0.0152 & 0.0137 & 0.0128  \\[6pt]
\multirow{1}{*}{Adversarial}   & $L_2$ & 1.7779 & - & - & - & 0.4016 \\
\bottomrule
\end{tabular*}
\end{table*}

\begin{figure}[ht]
    \centering
    
    \begin{subfigure}[b]{0.98\linewidth} 
        \centering
        \includegraphics[width=\linewidth,height=0.3cm,trim={0.12cm 0.5cm 0.1cm 0.5cm},clip]{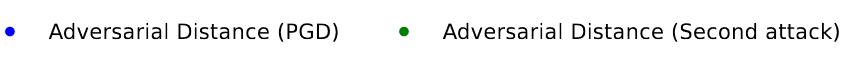} 
    \end{subfigure}
    \begin{subfigure}[b]{0.98\linewidth} 
        \centering
        \includegraphics[width=\linewidth,height=0.33cm,trim={0.1cm 0.5cm 0.1cm 0.5cm},clip]{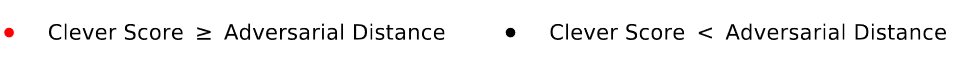} 
    \end{subfigure}
    
    \vspace{0.25cm} 

    \begin{subfigure}[t]{0.495\linewidth} 
        \centering
        \includegraphics[width=\linewidth,height=2cm,trim={0cm 0.4cm 0.3cm 0.3cm},clip]{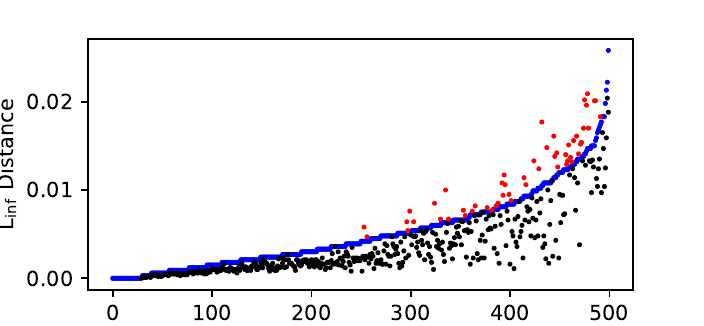}
        \vspace*{-5.5mm}
        \caption{Standard Model, $L_\infty$ norm}
        \label{fig:std-1024-500-1}
    \end{subfigure}
    \hfill
    \begin{subfigure}[t]{0.495\linewidth} 
        \centering
        \includegraphics[width=\linewidth,height=2cm,trim={0cm 0.4cm 0.3cm 0.3cm},clip]{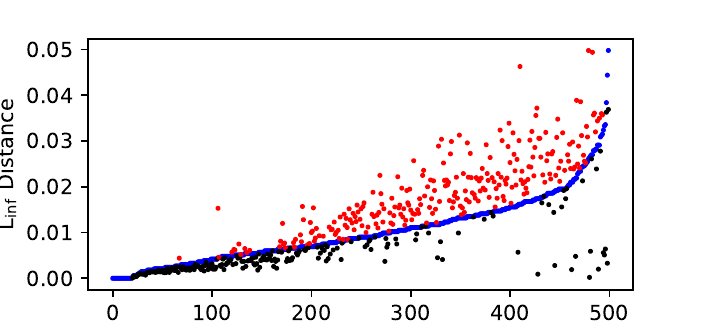}
        \vspace*{-5.5mm}
        \caption{Robust Model, $L_\infty$ norm}
        \label{fig:rob-1024-500-1}
    \end{subfigure}
    
    \vspace{0.25cm} 

    \begin{subfigure}[b]{0.495\linewidth} 
        \centering
        \includegraphics[width=\linewidth,height=2cm,trim={0.2cm 0.4cm 0.3cm 0.3cm},clip]{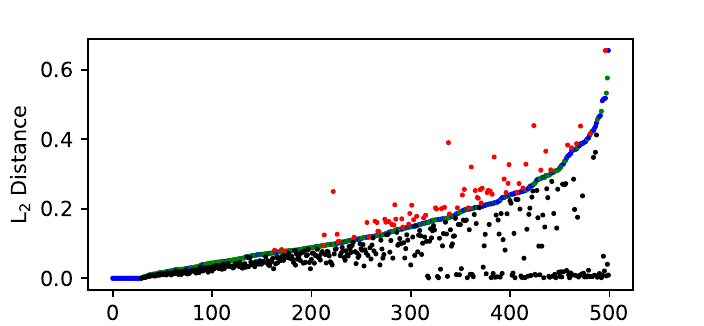}
        \vspace*{-5.5mm}
        \caption{Standard Model, $L_2$ norm}
        \label{fig:std-1024-500-2}
    \end{subfigure}
    \hfill
    \begin{subfigure}[b]{0.495\linewidth} 
        \centering
        \includegraphics[width=\linewidth,height=2cm,trim={0.1cm 0.4cm 0.3cm 0.3cm},clip]{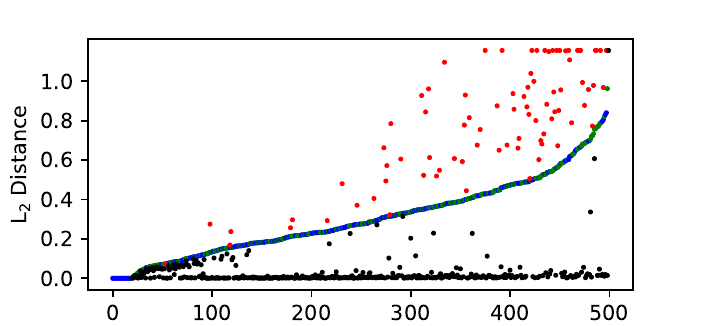}
        \vspace*{-5.5mm}
        \caption{Robust Model, $L_2$ norm}
        \label{fig:rob-1024-500-2}
    \end{subfigure}
    
    \vspace{0.25cm} 

    \begin{subfigure}[b]{0.495\linewidth} 
        \centering
        \includegraphics[width=\linewidth,height=2cm,trim={0.3cm 0.4cm 0.3cm 0.3cm},clip]{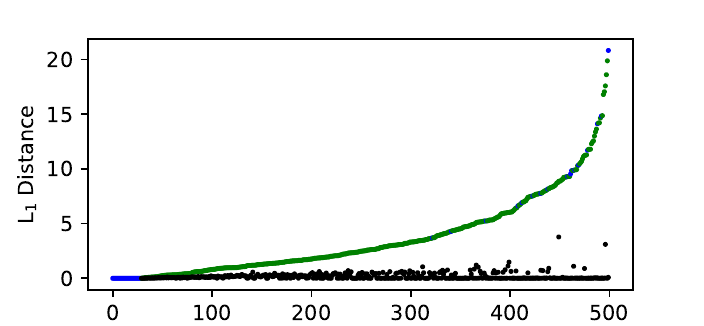}
        \vspace*{-5.5mm}
        \caption{Standard Model, $L_1$ norm}
        \label{fig:std-1024-500-inf}
    \end{subfigure}
    \hfill
    \begin{subfigure}[b]{0.495\linewidth} 
        \centering
        \includegraphics[width=\linewidth,height=2cm,trim={0.3cm 0.4cm 0.3cm 0.3cm},clip]{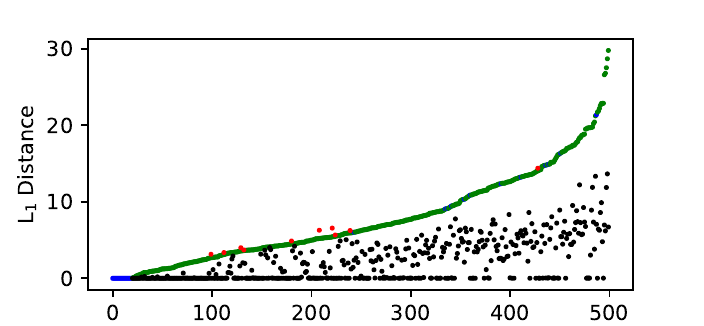}
        \vspace*{-5.5mm}
        \caption{Robust Model, $L_1$ norm}
        \label{fig:rob-1024-500-inf}
    \end{subfigure}
    
    \vspace{0.25cm} 
        
    \begin{subfigure}[b]{0.6\linewidth} 
        \centering
        \includegraphics[width=\linewidth,height=2.5cm,trim={0.3cm 0.4cm 0.3cm 0.3cm},clip]{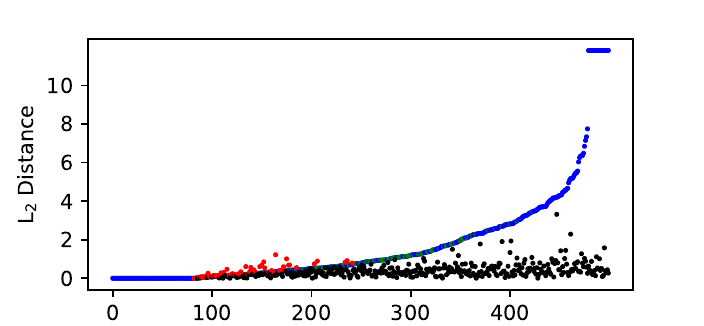}
        \vspace*{-5.5mm}
        \caption{Adversarial Model, $L_2$ norm}
        \label{fig:adv-1024-500-2}
    \end{subfigure}
        
    \caption{Adversarial distances for 500 images. For every image, the adversarial attack distance is shown in ascending order (0 for initially misclassified images, maximum of all adversarial distances for images where no adversarial example could be found). The color indicates whether Algorithm \eqref{alg:early_stopping} or the second attack was tighter. For all images correctly classified initially, CLEVER is also shown (1024/500 parameterisation). The color indicates whether it is a plausible or an incorrect lower bound.}
    \label{fig:adv_vs_cl}
\end{figure}

\subsection{CLEVER Score as a Lower Bound}

The figures \ref{fig:std-1024-500-1} to \ref{fig:adv-1024-500-2} show the comparison between CLEVER (for its most reliable 1024-500 parameter setup) and the minimal adversarial attack distance on 500 images, plotted for all 3 norms for the standard and robust models as well as for the adversarial model on $L_2$-norm. The images are sorted by adversarial distance as returned by Algorithm \eqref{alg:combined}. Misclassified points are assigned an adversarial distance of 0. Ideally, we expect the CLEVER scores to be just below the adversarial distance for most points. For the standard model and $L_1$ norm, it is clearly visible that CLEVER massively underestimates the adversarial distance, with many images having scores of 0. In contrast, for the robust model and $L_1$ norm, CLEVER seems to work better, although many images still have CLEVER scores of 0. For the standard model on $L_2$ and $L_\infty$ as well as the adversarial model on $L_2$, CLEVER gives reasonable estimates, with about 15\% of the CLEVER scores higher than their respective adversarial distances, indicating an incorrect lower bound estimate. For the robust model, CLEVER scores are unreliable on both norms. For $L_2$, about 18\% of the CLEVER scores are incorrect lower bounds and most of the rest are close to zero. For $L_\infty$, 54\% are incorrect lower bounds. 

In the plot for the adversarial model it is also visible how a larger ratio of points being misclassified as the models clean accuracy is lower. Also, 4\% of points cannot be succesfully attacked by our algorithm \ref{alg:combined}, in which case we assume those points to have the maximum adversarial distance found for any other point (see the top right corner of the diagram). An overview of all mean CLEVER scores for all parameter setups compared to the mean adversarial attack distance can be found in Table \ref{tab:adv_cl}.


\section{Discussion} \label{section:discussion}
Our results shed light on the effectiveness of a selection of evaluation methods for adversarial distance estimation. First, we emphasise that most of this evaluation was carried out with two models that were not trained to be adversarially robust. A comparison with the adversarial model show clearly that the latter is much more robust in terms of adversarial distance according to our method. 

In our experiments, we used implementations of attacks from only one popular adversarial robustness toolbox. It may be that another toolbox has already built an estimator like Algorithm \eqref{alg:early_stopping}.

In our experiments, we found our iterative attack algorithm with early stopping and a small step size, to be an effective baseline in terms of computational efficiency and adversarial distance estimates compared to a number of other adversarial attacks. Surprisingly, it is particularly effective compared to the ART implementations of FGSM \cite{goodfellow2014explaining} with early stopping and DeepFool \cite{moosavi2016deepfool}, both deliberately designed to estimate exactly this minimal adversarial distance.

We found that our estimation algorithm should be supported by a second algorithm that provides tight estimates of adversarial distances instead of using its entire attack budget. There are several reasons for this: 
\begin{enumerate}
    \item Models trained adversarially using one particular method, where that method may then be less effective at estimating adversarial distance. 
    \item Ensembles of adversarial attacks are state of the art for robustness evaluation \cite{croce2020reliable}.
    \item For all norms, but $L_1$ in particular, there exist attacks that appear to be more effective than PGD. The effectiveness of EAD on $L_1$ is probably due to its dual regularisation technique, which combines the sparsity of $L_{1}$ with the evenly distributed perturbations of the $L_{2}$ norm.
\end{enumerate}

In our experiments, we expected the CLEVER score to be an effective estimator of an adversarial distance lower bound. The visualisations of the results as in Figures \ref{fig:adv_vs_cl} allows us to discuss this expectation. It leads us to conclude that for our experiments, CLEVER is a rather unreliable estimator, even when parameterised to sample within the perfect norm distance derived from the previous attacks. This is true in particular when evaluating the robust model, probably because it was not trained to be strictly smooth using adversarial training or smoothing methods. However, as it was trained to be robust to random corruptions, it may have lulled CLEVER, which uses a random sampling scheme for estimation, into a false sense of security. In fact, the mechanism of a wrong estimation of CLEVER on a non-smooth model is explained in detail in \cite{goodfellow2018gradient}. Countering this phenomenon with many more samples is inefficient, but a different sampling scheme may help \cite{webb2018statistical}. For a reliable lower bound on adversarial distance, robustness verification methods with guarantees are needed. For the adversarial model, CLEVER worked more reliably, but we still found counter examples among the less robust points in particular. The precision of CLEVER may be improved by using the adversarial attack distance of each individual point as CLEVERs $\epsilon$ value in order to not waste any samples for its estimation.

We found from results such as in Table \ref{tab:adv_cl} that the mean adversarial distance captures the differences in adversarial robustness of all 3 models well, while the adversarial distance distributions across all points provide more details. On the other hand, for the adversarially trained model, our method has trouble finding an adversarial example on some of the points due to their high robustness. Nevertheless, we emphasise the usefulness of measuring the adversarial distance. For example, it helps evaluate the robustness of the robust model compared to the standard model, with its adversarial distance being twice as high. This is plausible, as research suggests a positive effect of random data augmentations such as those used for the robust model, on adversarial robustness \cite{ford2019adversarial}. However, a typical benchmark evaluation of $L_\infty$ adversarial accuracy with our PGD attack and an attack budget of $8/255$ would evaluate the standard model at 0\% and the robust model at 1.4\% adversarial accuracy, indicating little difference. Our evaluation gives a different and more informative impression about the adversarial robustness of the models compared to each other. We summarize that in line with the findings in \cite{Bastani16}, adversarial distance is a nuanced and useful measure of robustness at least for comparing models with about similar adversarial accuracy as in the example above.

\section{Conclusion}
This paper proposes a practical approach to estimating the (mean) adversarial distance of classifiers. We find that our simple algorithm provides a solid basis for this estimation, and propose a combination of several attacks and a certification methods to provide an overall assessment of a model's adversarial robustness. While our attack methods work effectively to estimate an upper bound on the adversarial distance for different norm distances, the CLEVER score certification does not provide reliable lower bounds in our experiments. We highlight the value of adversarial distance as a metric to consider for an overall robustness evaluation of a machine learning classifier. Future work should explore different (iterative) adversarial attacks to enable tight upper bounds on adversarial distance for various models and applications. Those results can in turn be used to evaluate the tightness of lower bounds on adversarial distance from verification approaches outside CLEVER score, or the validity of robustness certification approaches.

\bibliographystyle{asmeconf}  
\bibliography{Main-bib}

\end{document}